\documentclass{article} 
\PassOptionsToPackage{table,xcdraw}{xcolor}
\usepackage{iclr2025_conference,times}

\usepackage{amsmath,amsfonts,bm}

\def\eqref#1{equation~\ref{#1}}

\def\1{\bm{1}}

\DeclareMathAlphabet{\mathsfit}{\encodingdefault}{\sfdefault}{m}{sl}
\SetMathAlphabet{\mathsfit}{bold}{\encodingdefault}{\sfdefault}{bx}{n}

\usepackage{hyperref}
\usepackage{url}
\usepackage{bbm}
\usepackage{booktabs}
\usepackage{multirow}
\usepackage{graphicx}
\usepackage{kotex}
\usepackage{tabularx}
\usepackage{colortbl}
\usepackage{tcolorbox}
\usepackage{wrapfig}
\usepackage{algorithm}
\usepackage{algpseudocode}
\usepackage{amsmath}
\newcommand{\llama}{LLaMA}
\usepackage{enumitem}
\usepackage[skip=3pt]{caption}
\usepackage{calc}
\usepackage{varwidth}
\usepackage{float}
\title{Flex-TravelPlanner: A Benchmark for Flexible Planning with Language Agents}

\author{Juhyun Oh, Eunsu Kim \& Alice Oh  \\
School of Computing\\
KAIST\\
Daejeon, Republic of Korea \\
\texttt{\{411juhyun,kes0317\}@kaist.ac.kr, alice.oh@kaist.edu} \\
}

\iclrfinalcopy 
\begin{document}

\maketitle

\begin{abstract}

Real-world planning problems require constant adaptation to changing requirements and balancing of competing constraints.
However, current benchmarks for evaluating LLMs' planning capabilities primarily focus on static, single-turn scenarios.
We introduce Flex-TravelPlanner, a benchmark that evaluates language models' ability to reason flexibly in dynamic planning scenarios. Building on the TravelPlanner dataset~\citep{xie2024travelplanner}, we introduce two novel evaluation settings: (1) sequential constraint introduction across multiple turns, and (2) scenarios with explicitly prioritized competing constraints.
Our analysis of GPT-4o and Llama 3.1 70B reveals several key findings: models' performance on single-turn tasks poorly predicts their ability to adapt plans across multiple turns; constraint introduction order significantly affects performance; and models struggle with constraint prioritization, often incorrectly favoring newly introduced lower priority preferences over existing higher-priority constraints.
These findings highlight the importance of evaluating LLMs in more realistic, dynamic planning scenarios and suggest specific directions for improving model performance on complex planning tasks. The code and dataset for our framework are publicly available at \url{https://github.com/juhyunohh/FlexTravelBench}.

 
\end{abstract}

\section{Introduction}

Planning is a complex cognitive task, often requiring agents to adapt to changing circumstances and prioritize among competing goals.
Real-world planning problems, such as travel itinerary planning, rarely present all constraints upfront; instead, constraints are typically introduced and modified incrementally.
Recent benchmarks have made significant progress in evaluating LLMs' planning capabilities, but they primarily focus on static, one-shot scenarios rather than the dynamic, constraint-evolving nature of real-world planning~\citep{xie2024travelplanner, zheng2024natural, valmeekam2023planning}. 
While multi-turn evaluation approaches have been studied, benchmarks specifically tailored to planning tasks remain unexplored.
Existing work like MT-Eval~\citep{kwan-etal-2024-mt} focuses on conversational abilities, while studies investigating interactive problem-solving and reasoning ability (e.g., \citet{wangmint, kim2024llm}) rely on explicit feedback, unlike real-world scenarios where models must autonomously detect conflicts and determine appropriate plan revisions.

To address these limitations, we introduce Flex-TravelPlanner, a novel evaluation framework for assessing LLM flexible reasoning in dynamic, multi-turn planning. Building upon TravelPlanner~\citep{xie2024travelplanner}, our benchmark focuses on two key aspects: the ability to revise plans in response to incrementally changing constraints, and the capacity to effectively prioritize among constraints of varying importance. 
Through this framework, we investigate: 1) the impact of sequential vs. parallel constraint presentation on LLM performance; and 2) LLM's ability to leverage constraint priorities when full constraint satisfaction is infeasible.

Evaluating GPT-4o and LLaMA 3.1 70B in zero-shot settings, we find: 1) Strong single-turn performance does not guarantee robust multi-turn performance; 
2) Constraint order matters, with both models showing higher success rates when constraints requiring consideration of the entire itinerary (e.g., budget) are introduced after constraints affecting individual choices (e.g., hotel room type); and 3) Models struggle with constraint prioritization, often incorrectly favoring newly introduced lower priority preferences over existing constraints.
These findings highlight critical areas for future research in enhancing LLM planning capabilities, particularly in dynamic and prioritized constraint scenarios, paving the way for more robust real-world applications.

\section{Flex-TravelPlanner}

\subsection{Flexible Planning Evaluation Framework}
We introduce a novel evaluation framework designed to assess the flexible reasoning abilities of language agents in dynamic, multi-turn planning scenarios. This framework focuses on evaluating how well agents can adapt their plans as new requirements or changes to existing requirements are introduced over multiple interactions. Specifically, it addresses the challenges of constraint addition and revision, mirroring the dynamic nature of real-world planning.

\begin{figure}[htbp]
    \centering        
    \includegraphics[width=0.9\columnwidth]{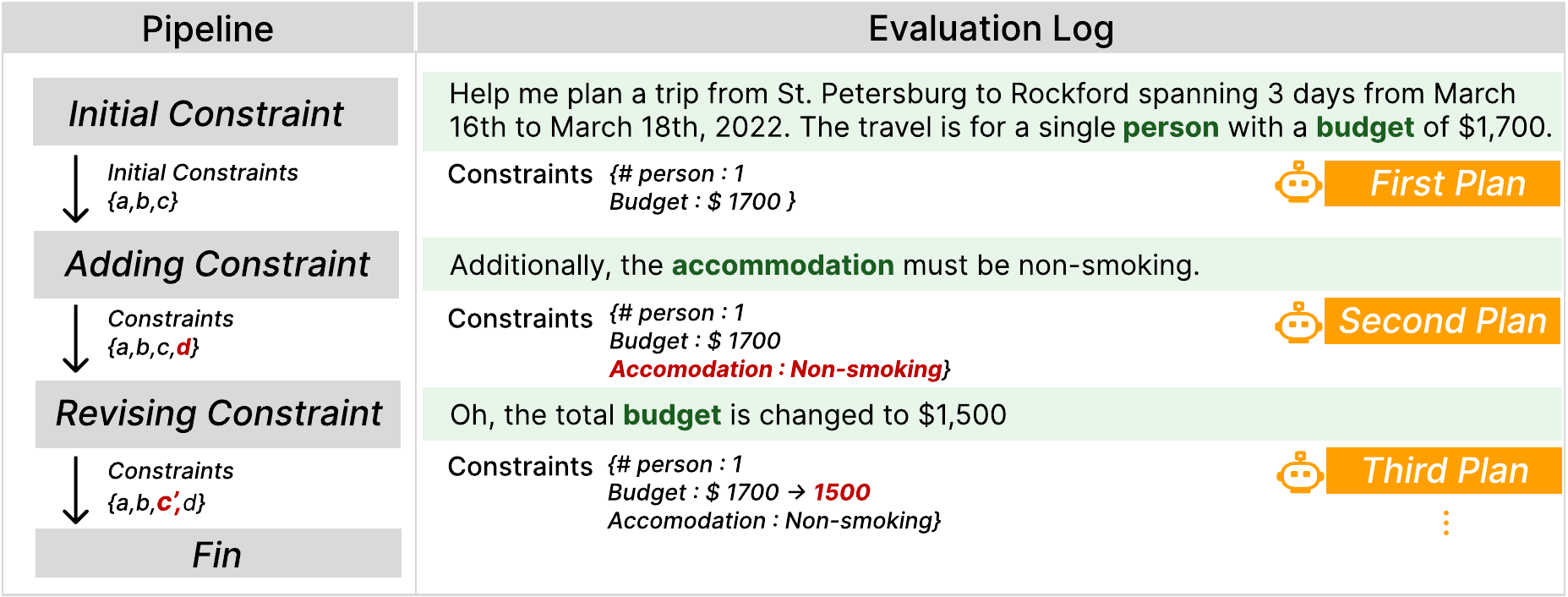}
        \caption{Framework for evaluating flexible planning in Language agents. The right panel demonstrates a travel planning example with evolving budget and accommodation constraints.}
        \label{fig:overview}
\end{figure} 

Our framework evaluates the agent's ability to:
\begin{enumerate}[noitemsep,topsep=0pt,left=10pt]
  \item \textbf{Adapt to New Constraints:} When constraints are added or revised in subsequent turns, can the agent update its plan to incorporate these changes while maintaining overall plan validity and satisfaction of existing constraints?
  \item \textbf{Prioritize Among Conflicting Constraints:} When new constraints conflict with existing ones, can the agent prioritize among them, making appropriate trade-offs and focusing on satisfying the most critical requirements?
\end{enumerate}


We specifically design Flex-TravelPlanner dataset, described in the following section, to evaluate these two core aspects of flexible reasoning within the context of travel planning. 



\subsection{Flex-TravelPlanner Dataset}
Flex-TravelPlanner evaluates LLMs' planning skills in dynamic, multi-turn scenarios. While leveraging data from the TravelPlanner~\citep{xie2024travelplanner} dataset, our benchmark focuses specifically on evaluating flexible reasoning with incrementally changing and prioritized constraints. 
All necessary reference information (e.g., restaurant details) is provided alongside each query to enable direct evaluation of the models' planning capabilities.


\noindent\textbf{Constraints.}
We utilize a subset of the hard constraints defined in TravelPlanner (see Table \ref{tab:constraint_introduction} for details). These constraints represent personalized user needs, such as budget and room rules. 
We add restaurant ratings as a new constraint to build the priority-aware plan revision dataset.
We classify constraints as \textit{global}, impacting the entire itinerary (e.g., budget), or \textit{local}, applying to specific elements (e.g., hotel room type).
This distinction is important for understanding how LLMs handle different scopes of planning requirements.

\begin{table*}[t]
\centering
\resizebox{\textwidth}{!}{
\begin{tabular}{lll}  
\toprule
\rowcolor[HTML]{D9D9D9} \textbf{Type} & \textbf{Constraint} & \textbf{Description} \\  
\midrule

\textsc{Global} & Budget & The total budget of the trip. \\ 
\cmidrule(lr){1-3} 

\textsc{Local} & Room Rule & Include “No parties”, “No smoking”, 
“No children under 10”, “No pets”, and “No visitors”. \\

& Room Type & Include “Entire Room”, “Private Room”, 
“Shared Room”, and “No Shared Room”. \\

& Cuisine & Include “Chinese”, “American”, “Italian”, 
“Mexican”, “Indian”, “Mediterranean”, and “French”. \\

& Rating & Minimum required rating of the restaurants. \\

\bottomrule
\end{tabular}%
}
\caption{Constraint description. Type indicates the scope of the constraint.}
\label{tab:constraint_introduction}
\vspace{-3mm}
\end{table*}
\noindent\textbf{Constraint-Adaptive Plan Revision.}
To assess constraint-adaptive plan revision, we construct multi-turn scenarios using 120 queries from TravelPlanner's validation set that include both global and local constraints. Using the same set of constraints per query, we compare three introduction patterns: all-at-once (N), 2-turn (N-1, 1), and 3-turn (N-2, 1, 1) scenarios. Each query includes 1-3 local constraints for groups of 2-8 people. 
Figure~\ref{fig:overview} illustrates a sample query with new constraints. To ensure that all plans are solvable, we only test constraint addition, and not revision for this study. 
\noindent\textbf{Priority-Aware Plan Revision.}
We evaluate priority-aware plan revision by introducing potentially conflicting soft and hard constraints. Soft constraints include cuisine preferences (``try \textit{cuisine type} \textit{N} times if possible'') and rating preferences (``visit restaurants rated \textit{minimum rating} or higher \textit{M} times if possible''), which may conflict with the budget constraint. We test the LLM's constraint prioritization using 134 scenarios. See Appendix~\ref{appendix:priority} for dataset construction details.

\noindent\textbf{Evaluation Metric.}
We evaluate constraint satisfaction using Constraint Pass Rate, which measures the ratio of passed constraints to total constraints across all plans:
\begin{equation}
\text{Constraint Pass Rate} = \frac{\sum_{p \in P}\sum_{c \in C_p} \mathbbm{1}_{\text{passed}(c, p)}}{\sum_{p \in P} |C_p|},
\end{equation}
where $P$ is the set of plans, $C_{p}$ is the set of constraints for plan $p$, and $\text{passed}(X,Y)$ indicates whether plan $Y$ satisfies constraint $X$.

\section{Experiments and Results}

\subsection{Experimental Settings}
We test two models, representing most popular choices in proprietary and open-source LLMs: GPT-4o \citep{dubey2024llama}\footnote{We use GPT-4o-0514 via OpenAI API.} and \llama 3.1 70B \citep{achiam2023gpt}\footnote{We use meta-llama/Meta-Llama-3.1-70B-Instruct-Turbo via TogetherAI API}. We use the same set of 120 queries for the multi-turn constraint-adaptive plan revision experiments, testing each query in 1-turn, 2-turn, and 3-turn scenarios. For the priority-aware plan revision experiments, we use a separate set of 134 queries, and test in 2-turn scenarios.

All experiments are conducted in a zero-shot setting, evaluating the models' direct performance without any prompting or fine-tuning. While we do not explicitly prevent models from using chain-of-thought reasoning before planning, we assess their performance based on the final plan generated. In all our settings, all previous turns are provided as history. Constraints and reference information are provided in a structured JSON format. Details of the plan format is in the Appendix~\ref{appendix}.

\subsection{Results}
\begin{wrapfigure}{r}{0.55\textwidth}
    \vspace{-6mm}
    \centering
    \includegraphics[width=0.95\linewidth]{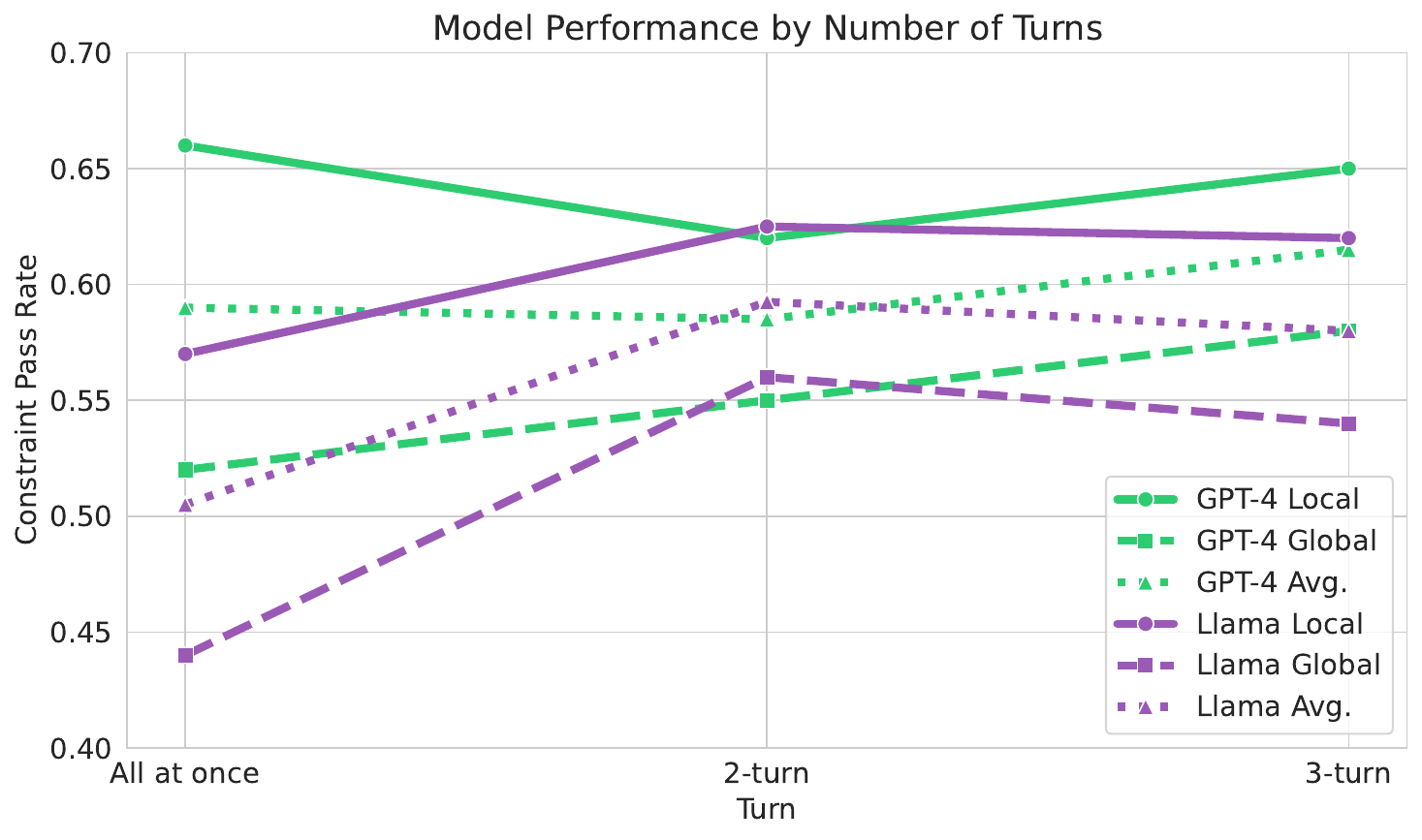}
    \caption{Local, global and average pass rates of the two models across turns.}
    \label{fig:main_figure}
    \vspace{-8.0mm}
\end{wrapfigure}

\paragraph{Constraint-Adaptive Plan Revision.}

Figure \ref{fig:main_figure} illustrates the constraint pass rates for both models across 1-, 2-, and 3-turn scenarios. While GPT-4o demonstrates higher pass rates than LLaMA when constraints are presented all at once, LLaMA outperforms GPT-4o in both global and local constraints during 2-turn interactions. Unlike GPT-4o, LLaMA shows improved performance when constraints are introduced sequentially in multi-turn settings rather than all at once, suggesting that strong single-turn performance doesn't necessarily translate to robust multi-turn capabilities.
The improved performance in multi-turn scenarios compared to single-turn interactions suggests sequential constraint introduction as an effective strategy for complex planning tasks.

\begin{figure*}[htbp]
    \centering
    
    \begin{minipage}{0.48\columnwidth}
        \centering
        \small
        \begin{tabular}{@{}l@{\hspace{6pt}}cc@{\hspace{12pt}}cc@{}}
        \toprule
        & \multicolumn{2}{c}{GPT-4o} & \multicolumn{2}{c}{Llama 3.1 70B} \\ 
        \cmidrule(r){2-3} \cmidrule(l){4-5}
        & L & G & L & G \\ 
        \midrule
        1-turn \\
        \hspace{0.3mm}(All at once) & 0.66 & 0.52 & 0.57 & 0.44 \\ 
        \midrule
        2-turn \\
        \hspace{3mm}\textbf{+G} & 0.62 & \textbf{0.63} & 0.62 & \textbf{0.67} \\
        \hspace{3mm}+L & 0.62 & 0.47 & 0.63 & 0.45 \\ 
        \midrule
        3-turn \\
        \hspace{3mm}+L \textbf{+G} & 0.65 & \textbf{0.62} & 0.63 & \textbf{0.56} \\
        \hspace{3mm}+G +L & 0.65 & 0.54 & 0.61 & 0.52 \\
        \bottomrule
        \end{tabular}
        \caption{Local (L), Global (G), and Average (Avg.) constraint pass rates across conditions.}
        \label{tab:main_constraint_addition}
    \end{minipage}%
    \hfill
    \begin{minipage}{0.5\columnwidth}
        \centering
        \includegraphics[width=0.9\columnwidth]{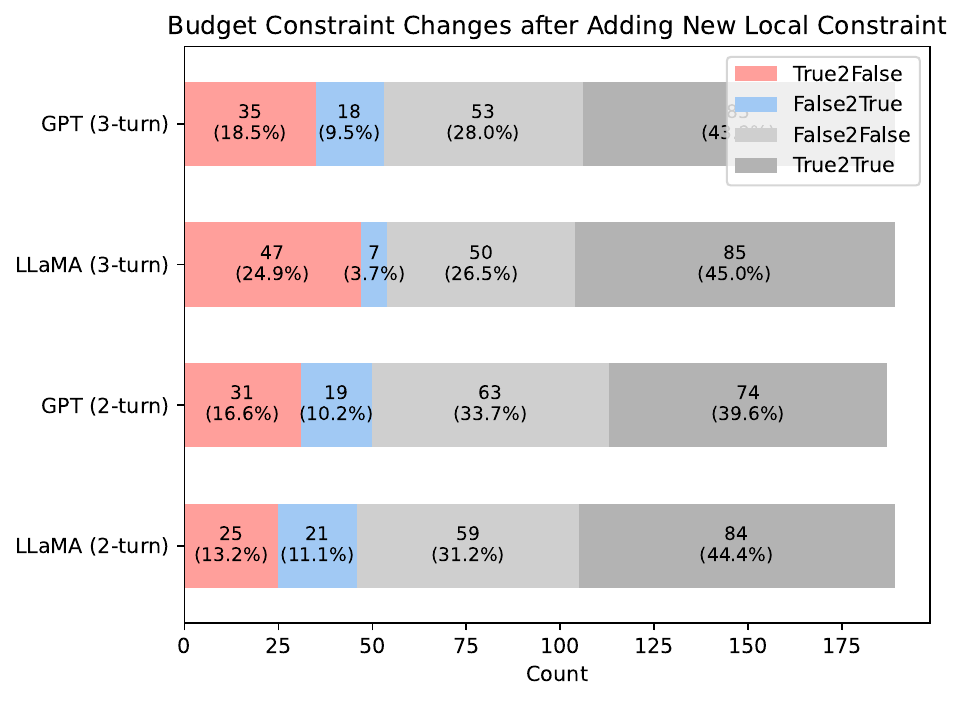}
        \vspace{-1.4mm}
        \caption{Changes in Budget pass rates when local constraints are added.}
        
        \label{fig:track_change_by_turns}
    \end{minipage}
    \vspace{-3mm}
\end{figure*}

The order of constraint introduction significantly impacts model performance, particularly for global constraints. When global constraints (e.g., budget requirements) are introduced before local constraints, models struggle to maintain compliance as new constraints are added.
Figure \ref{tab:main_constraint_addition} demonstrates this through constraint pass rates across different introduction orders. While local constraint performance remains stable, global constraint performance varies substantially based on sequence. Both models achieve higher pass rates when budget constraint (global) are introduced later -- LLaMA improves from 0.52 to 0.67, and GPT-4o from 0.47 to 0.63 in 2-turn scenarios.

\begin{wrapfigure}{r}{0.5\textwidth}
    \centering
    \includegraphics[width=0.5\textwidth]{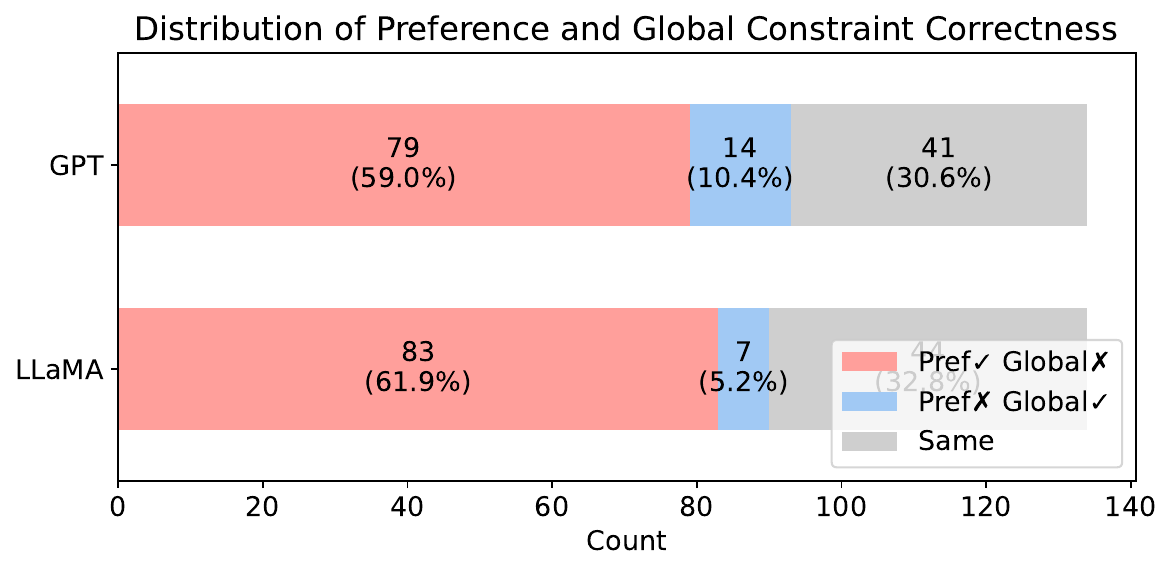}
    \caption{Preference and global constraint pass rate after conflicting preference constraint introduction in the second turn.}
    \label{fig:preference_stack}
\end{wrapfigure}
Figure \ref{fig:track_change_by_turns} tracks budget constraint satisfaction changes when local constraints are subsequently introduced. Plans initially satisfying the budget constraint often become non-compliant after local constraints are added, with True-to-False (red bars) transitions consistently outnumbering False-to-True (blue bars) corrections. This suggests that models struggle to maintain global constraint compliance while accommodating new local requirements, rather than using additional turns as opportunities to fix non-compliant plans.

\noindent\textbf{Priority-Aware Plan Revision.}
Figure \ref{fig:preference_stack} illustrates how LLMs handle conflicts between hard, global constraints (budget) and preference constraints. While LLMs should prioritize satisfying budget constraints even when they conflict with newly introduced preferences (blue bars), both models frequently violate budget constraints to accommodate preferences -- 59\% for GPT and 61.9\% for LLaMA. This indicates that current LLMs struggle to properly prioritize between hard constraints and low-priority preferences in their planning process.

\section{Conclusion}
This work introduces Flex-TravelPlanner, a novel benchmark for evaluating LLM performance in dynamic, multi-turn planning scenarios with prioritized constraints. Our findings reveal a disconnect between single-turn and multi-turn performance, and that constraint introduction order significantly impacts plan quality—models particularly struggle to maintain global constraints when local constraints are subsequently introduced. While sequential constraint introduction shows promise as a strategy for complex planning tasks, both GPT-4o and LLaMA exhibit fundamental weaknesses in maintaining constraint hierarchies, highlighting critical challenges for developing reliable LLM-based planning systems for real-world applications.

\bibliography{iclr2025_conference}
\bibliographystyle{iclr2025_conference}

\appendix
\newpage
\section{Appendix}
\subsection{Question Generation - Priority-Aware Plan Revision}
\label{appendix:priority}

The questions are divided into two types: cuisine and rating. Questions are generated based on predefined formats, where variables within the format (e.g., cuisine type, rating, etc.) and budget are modified. The formats used and question generation methods are as follows:

\paragraph{Cuisine Question Format}
``I prefer to try a \{cuisine type\} restaurant at least \{N\} times if possible.''

To create conflicts with the budget in this type,
for a chosen cuisine type (e.g., Asian cuisine) and N, we first calculate the cost of visiting the lowest-priced restaurants of that cuisine type N times. Then, we set a budget that is lower than what would be required to satisfy all other local constraints at their minimum options while meeting this cuisine preference. This deliberately creates a conflict between the query and the budget constraint.

\paragraph{Rating Question Format}
``I prefer to visit restaurants with a minimum rating of \{minimum rating\} at least \{M\} times if possible.''

To create conflicts with the budget in this type,
for a chosen minimum rating value (e.g., 4.0) and N, we first calculate the cost of visiting the lowest-priced restaurants meeting this rating threshold N times. Then, we set a budget that is lower than what would be required to satisfy all other local constraints at their minimum options while meeting this rating preference. This deliberately creates a conflict between the query and the budget constraint.
\newpage

\subsection{Prompts}
Following is the prompt template used for generating plans. Models are given the reference information required, example plan format, and the query. 
\begin{tcolorbox}
\small
[Reference information]:\\
All costs are per one person, one night.\\
\{reference\_data\}\\

[Plan Format]: [\{ \\
        ``days": 1, \\
        ``current\_city": ``from Dallas to Peoria", \\
        ``transportation": ``Flight Number: 4044830, from Dallas to Peoria, Departure Time: 13:10, Arrival Time: 15:01",\\
        ``breakfast'': ``-",\\
        ``attraction": ``Peoria Historical Society, Peoria;Peoria Holocaust Memorial, Peoria;",\\
        ``lunch": ``-",\\
        ``dinner": ``Tandoor Ka Zaika, Peoria",\\
        ``accommodation": ``Bushwick Music Mansion, Peoria"\\
    \},\\
    \{\\
        ``days": 2,\\
        ``current\_city": ``Peoria",\\
        ``transportation": ``-",\\
        ``breakfast": ``Tandoor Ka Zaika, Peoria",\\
        ``attraction": ``Peoria Riverfront Park, Peoria;The Peoria PlayHouse, Peoria;Glen Oak Park, Peoria;",\\
        ``lunch": ``Cafe Hashtag LoL, Peoria",\\
        ``dinner": ``The Curzon Room - Maidens Hotel, Peoria",\\
        ``accommodation": ``Bushwick Music Mansion, Peoria"\\
    \},\\
    \{\\
        ``days": 3,\\
        ``current\_city": ``from Peoria to Dallas",\\
        ``transportation": ``Flight Number: 4045904, from Peoria to Dallas, Departure Time: 07:09, Arrival Time: 09:20",\\
        ``breakfast": ``-",\\
        ``attraction": ``-",\\
        ``lunch": ``-",\\
        ``dinner": ``-",\\
        ``accommodation": ``-"\\
    \}]
\\
\\
\{question\}\\
Please refer to the given reference information only.\\

[Plan]:
\end{tcolorbox}
\label{appendix}

\end{document}